\documentclass{article} 
\usepackage{nips14submit_e,times}
\usepackage{hyperref}
\usepackage{graphicx}
\usepackage{url}
\usepackage[font=small,labelfont=bf]{caption}

\newcommand{\fig}[1]{Fig.~\ref{fig:#1}}
\newcommand{\tab}[1]{Table~\ref{tab:#1}}

\def\BE{\vspace{-0.0mm}\begin{equation}}
\def\EE{\vspace{-0.0mm}\end{equation}}
\def\BEA{\vspace{-0.0mm}\begin{eqnarray}}
\def\EEA{\vspace{-0.0mm}\end{eqnarray}}

\def\etal{{\textit{et~al.~}}}

\title{Deep Poselets for Human Detection}

\author{
Lubomir Bourdev$^{1}$
\And
Fei Yang$^{1}$
\And
Rob Fergus$^{1,2}$\\
\AND
$^1$Facebook AI Research\\
\And
$^2$New York University\\
\AND
{\tt\small \{lubomir, yangfei\}@fb.com}
\And
{\tt\small fergus@cs.nyu.edu}
}

\nipsfinalcopy 

\begin{document}

\maketitle

\begin{abstract}
We address the problem of detecting people in natural scenes using a part approach based on poselets. 
We propose a bootstrapping method that allows us to collect millions of weakly labeled examples for each poselet type. We use these examples to train a Convolutional Neural Net to discriminate different poselet types and separate them from the background class. We then use the trained CNN as a way to represent poselet patches with a Pose Discriminative Feature (PDF) vector -- a compact 256-dimensional feature vector that is effective at discriminating pose from appearance. We train the poselet model on top of PDF features and combine them with object-level CNNs for detection and bounding box prediction. The resulting model leads to state-of-the-art performance for human detection on the PASCAL datasets.
\end{abstract}

\section{Introduction}

Detecting humans in natural scenes is challenging due to the huge
variability in appearance, pose and occlusion. The recent state-of-the-art approaches of Girschik \etal \cite{Ross} and Sermanet \etal \cite{Sermanet13} differ mostly based on the algorithms they use to select candidate locations -- sliding windows or bottom-up segmentation.

The R-CNN
work of Girschik \etal \cite{Ross} shows how localization can be
performed by first proposing possible object locations and then
presenting each in turn to a convolutional network classifier. Although this
approach is somewhat slow, it currently has the leading performance
Imagenet and PASCAL detection tasks. 

The OverFeat system of Sermanet \etal \cite{Sermanet13} also uses a
convolutional network, but scans it over the entire image, rather than
a small set of locations. The top part of the network directly
predicts the coordinates of the bounding box. The proposed bounding
boxes are agglomerated into distinct object detections, which yields very
good performance on the Imagenet detection task. 

\begin{figure}[h!]
\begin{center}
\includegraphics[width=5in]{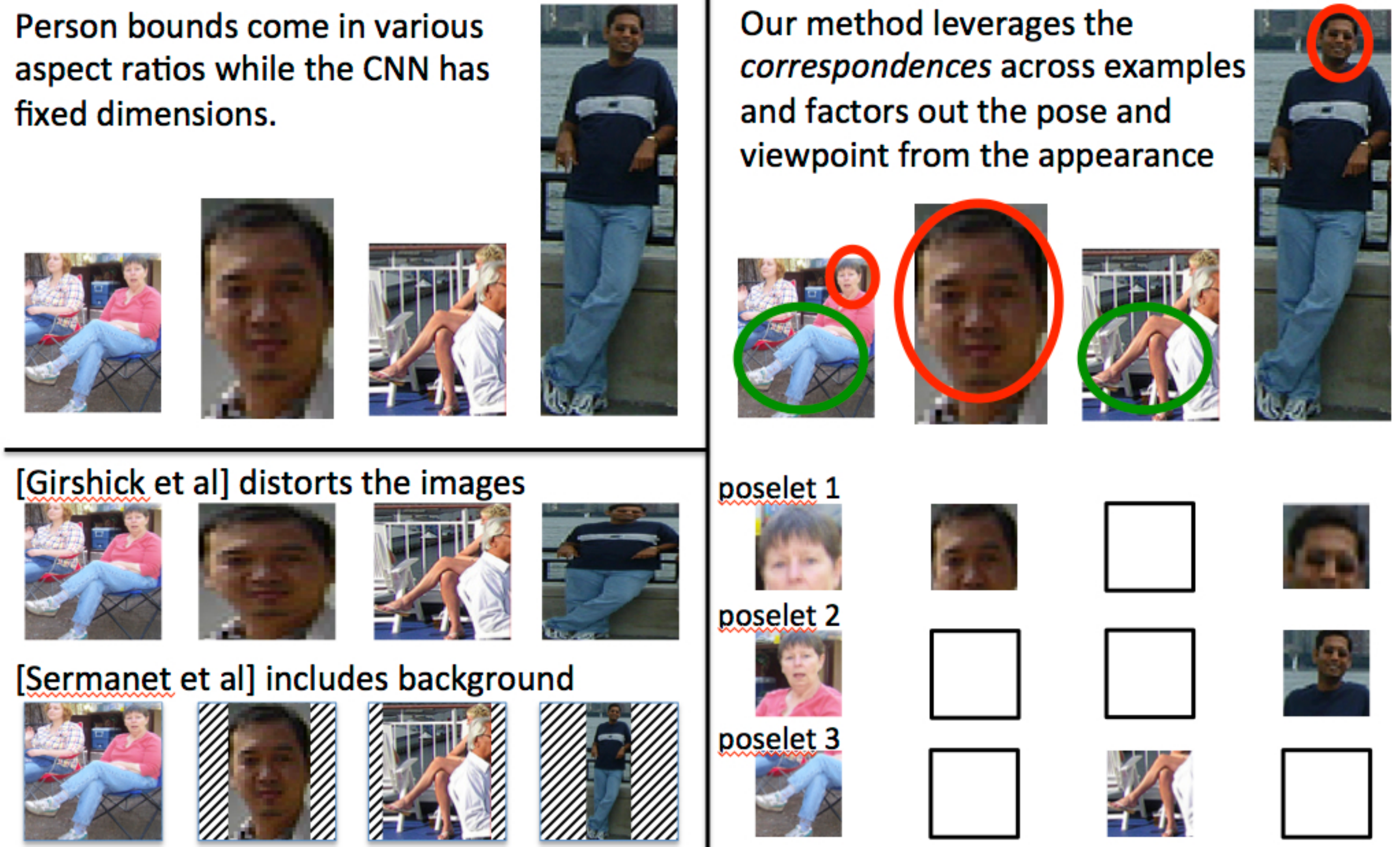}
\end{center}
\vspace*{0cm}
\caption{\textbf{Left:} The two state-of-the-art methods handle differently the problem of variable aspect ratios to fit them into the fixed-dimensional CNN. R-CNN's distorting the data makes the problem harder as the system has to learn everything under various degrees of distortions that don't occur in the natural world. OverFeat's solution doesn't distort the data but introduces sometimes lots of noisy background. Neither methods leverage the natural correspondences of parts.~\textbf{Right:} Our method doesn't distort the image and doesn't include unnecessary background. By factoring the pose and viewpoint from the appearance, poselets significantly simplify the variability of the positive class for each part and makes training part classifiers easier. Furthermore, poselets leverage the semantic correspondences in the input images, which which makes it easier to do high-level semantic reasoning. }
\vspace{-5mm}
\label{fig:regions}
\vspace*{0cm}
\end{figure}


One of the challenging aspects of person detection is that the visible portion of the person can have an arbitrary aspect ratio and it is hard to fit into the fixed dimensional input of the convolutional neural net. OverFeat and R-CNN take different approaches. In the case of OverFeat, the square scanning window approach can result in much background being included, and other times fitting only part of the person in the scanning window. To detect people, the holistic system of OverFeat has to learn to detect a huge number of patterns -- arbitrary parts of a person under arbitrary viewpoints, poses and appearances, which is particularly challenging because it has to learn to distinguish each from arbitrary non-person patterns. In the case of R-CNN the region proposals come from bottom-up segmentation and can have different aspect ratios. To fit them into the fixed size dimensions, the method resizes them, which sometimes results in very distorted patterns.  So the training examples for R-CNN are made more complex by including patterns that do not exist in the natural world, which makes it harder for the system to learn the correspondences across instances of the same object. Since convolutional nets are not aspect-ratio invariant, R-CNN would have to include plenty of examples with each possible aspect ratio to compensate for it. Furthermore, bottom-up methods are constrained by the ability of the segmentation system to detect suitable regions for the target object. If the segmentation fails to propose a suitable candidate region for a person, the CNN will not be given the opportunity to classify it.



\begin{figure}[h!]
\begin{center}
\includegraphics[width=1.7in]{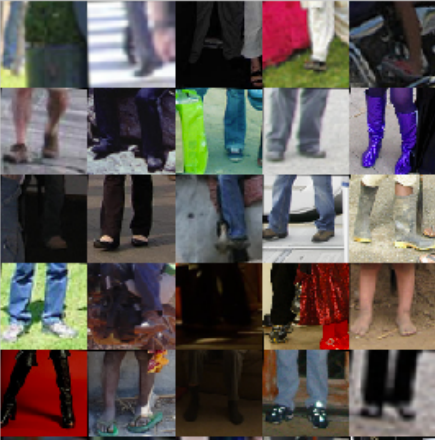}
\includegraphics[width=1.7in]{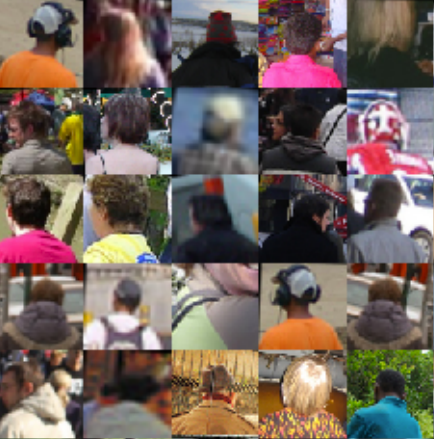}
\includegraphics[width=1.7in]{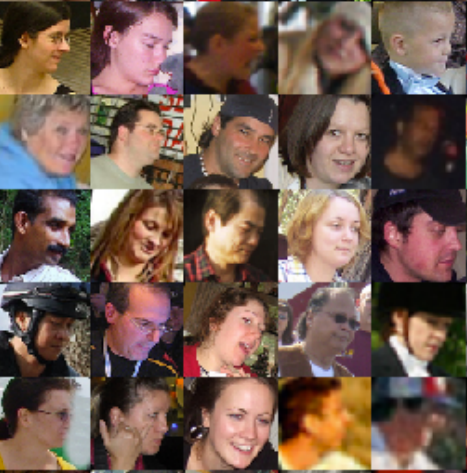}
\end{center}
\vspace*{-0.3cm}
\caption{Examples of poselet detections: legs of a front-facing person, back-facing head-and-shoulders, and a semi-profile face.}
\label{fig:teaser}
\vspace*{-0.3cm}
\end{figure}

We propose a third approach for detection -- a system that does not have to train with distorted aspect ratios or handle a large variability of positive examples within the same classifier.  Our method is inspired by the {\em Poselets} approach of Bourdev \etal \cite{Bourdev09}. Poselets are parts trained to respond to parts of a person under specific viewpoint and pose. Examples of poselets are shown in \fig{teaser}. Instances of each poselet type are detected in the test image, each of which votes on the location of the person. Poselet detections with compatible votes are clustered together into a person hypothesis. By constraining the training set of each poselet to examples of the same part of the body under the same viewpoint, the variability of the positive class of each poselet classifier is significantly reduced, training is easier, and the aspect ratio need not change. These benefits allowed poselets to maintain state-of-the-art results on the PASCAL person detection tasks until very recently before being overtaken by convolutional neural net methods, which, as numerous recent works demonstrate, work much better than the HOG features traditionally used by poselets.

We propose a deep learning version of the poselet architecture. CNNs require lots of training examples to avoid overfitting but building a large set of training examples for poselets is expensive. \emph{Our first contribution is a bootstrapping scheme} where we use the traditional poselets to collect lots of examples, automatically provide weak labels for them, and use this much larger training set to train deep poselet classifiers. Pose combinations have a heavy-tailed distribution. Lots of patterns are rare and it is hard to collect enough training examples to train a deep classifier for them. Furthermore, different vision tasks (detection, pose estimation, attribute extraction, segmentation) may require different set of optimal poselets. Training a deep poselet for each of them is time consuming. \emph{Our second contribution is using the deep representation to provide a generic pose-discriminative compact feature vector} of only 256 dimensions. We show that this feature vector allows us to quickly, and with very few examples, train an effective poselet classifier that the original HOG-based method was unable to do. Our method achieves state-of-the-art results on the challenging PASCAL datasets for the person detection category.

\subsection{Related work}

Notable early approaches to object detection focused on faces~\cite{Viola2001} and pedestrians~\cite{DalalTriggs}. A natural approach is to find individual parts of the person and then combine them into an overall detection. An early example is the constellation model of~\cite{Fergus03} whose parts location and appearance are jointly trained via EM. The implicit shape model of~\cite{leibe2004combined} employs a variable number of parts that use Hough voting to suggest the object location. A more recent popular example is the deformable parts
model by Felzenszwalb \etal \cite{dpm}, which consists of a small set of
part detectors plus a relative geometry term. Different views are
handled as components in a mixture model. Given bounding boxes in
training, the part appearances can be learned using a latent SVM
framework. However, even with several mixture components, the number
of distinct parts in the model is relatively small, i.e.~ a few dozen. Other notable approaches include: Filder \etal \cite{Fidler13}, Uijlings \etal \cite{Uijlings}, Dollar \etal \cite{dollar2010fastest}, and the Regionlets approach of Wang \etal \cite{wang}.

Lecun \etal introduced convnets \cite{lecun} for hand-writing recognition, but they have recently been shown to be very effective across a range of vision tasks -- classification~\cite{Krizhevsky12}, detection~\cite{Szegedy13,Sermanet13,sermanet-cvpr-13}, scene labeling~\cite{Farabet12}, face recognition~\cite{taigman14}, and attributes~\cite{panda} to name a few. Several works \cite{Ross,Josef,Matt}
have shown how deep networks can learn features that are far more
powerful than hand-crafted features such as HOG or SIFT.

\section{Approach}

Our method consists of four stages: (1) Training the deep pose representation, (2) training poselets on top of the deep features, (3) training object classifiers and (4) evaluating at test time.

\subsection{Training the Deep Pose Representation}

\begin{figure}[h!]
\begin{center}
\includegraphics[width=4.5in]{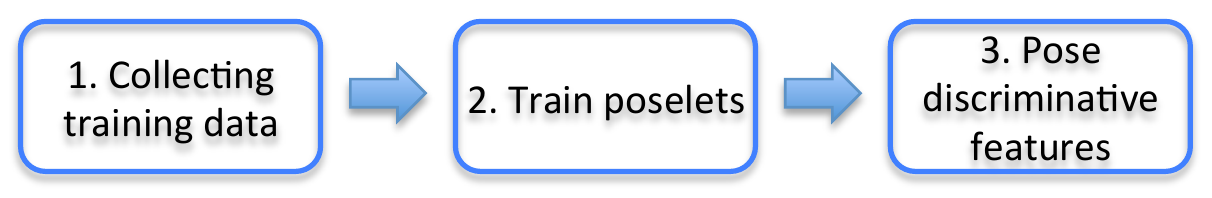}
\end{center}
\vspace*{-0.3cm}
\caption{Training our deep pose-discriminative features}
\label{fig:flowchart}
\vspace*{-0.3cm}
\end{figure}

\subsubsection{Collect Training Data}
The original poselets work \cite{Bourdev09} used SVMs trained on HOG
descriptors as poselet classifiers. In this paper, encouraged by the
recent successes of convolutional networks (convnets) for object recognition
tasks, we explore convnets for poselet classification. A major
obstacle is the huge amount of data needed to train large
convnets. As described in the next section, poselets require the training examples to be annotated with keypoints (joints, eyes, nose, etc), which is expensive to do at scale.

\begin{figure}[h!]
\begin{center}
\includegraphics[width=1.7in]{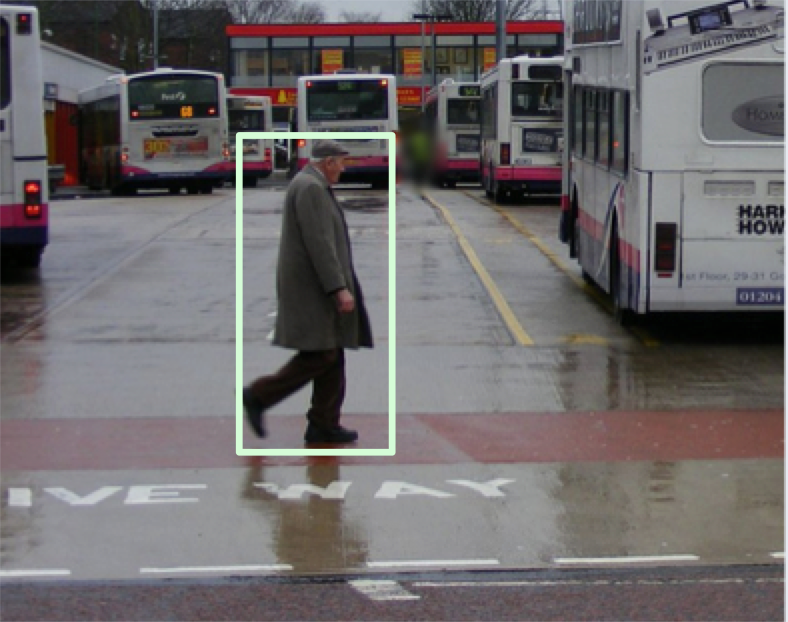}
\includegraphics[width=1.7in]{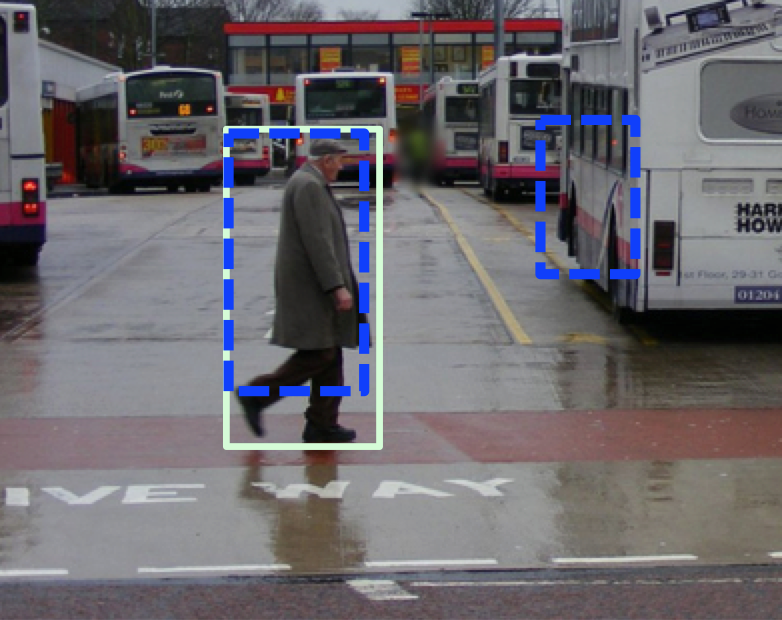}
\includegraphics[width=1.7in]{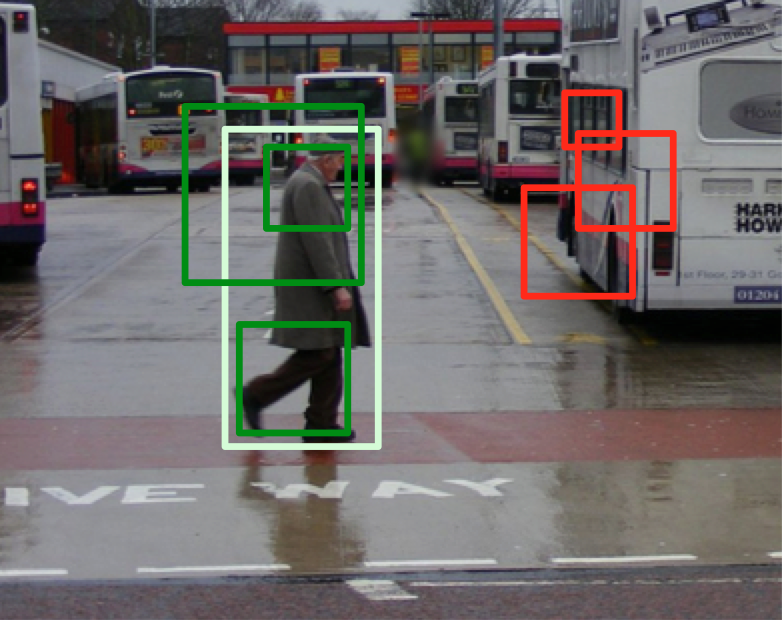}
\end{center}
\vspace*{-0.3cm}
\caption{Bootstrapping to collect more training data.~\textbf{Left:} We start with images with ground truth bounding boxes for people. \textbf{Middle:} We run the open-source poselet detectors and predict person hypothesis bounding boxes (shown in blue). We determine the ones whose intersection over union with ground truth is greater than 0.5 to be true positive. \textbf{Right:} All poselet activations supporting a true positive hypothesis are weakly-labeled as true positives (shown in green) and the rest as negative (shown in red).}
\label{fig:bootstrapping}
\vspace*{-0.3cm}
\end{figure}

We collect a large set of training examples by running the publicly available HOG-style poselet classifiers of~\cite{Bourdev09} on a large set of images on which we have only annotated the locations of people (see~\fig{bootstrapping}). Each person hypothesis suggested by the poselets is matched against the bounds of real people and hypotheses that overlap by more than 0.5 intersection-over-union are considered true positives. All poselet activations that support a true positive person hypothesis are considered true positive activations, and the ones not backed by a true positive are considered false positive activations. This bootstrapping mechanism allowed us, without significant annotation effort, to collect 20K positive training examples for each of the 150 poselet types of~\cite{Bourdev09} and we also collected 3 million false positive examples. The negative examples are "hard negative" as they are not just random patches but activations if a poselet not supported by a ground truth.

\subsubsection{Train Deep Net}

We extracted RGB patches of 61x61 dimensions associated with each poselet activation, subtracted the mean and used this data to train the convnet model shown in \tab{arch}. Since the poselets are
reasonably aligned and we want good localization, the architecture has
only a small amount of pooling, compared to convnets for whole image classification (e.g.\cite{Krizhevsky12}).

\begin{table}[h]
\begin{center}
    \begin{tabular}{| l | c |c | c |c | c |c| }
    \hline
Layer                     &   1                    &   2               &   3            &  4           &  5        &   6      \\  \hline
Stage                     &   conv+max     &  conv           &  conv       & conv      &  full     & full      \\  \hline
\#channels             &   64                 & 256              &  128         & 128       & 256     & 151     \\
Filter size               &   5 x 5             &  5 x 5           &  3 x 3        &  3 x 3    &  -           &  -        \\
Conv stride            &   2 x 2             & 1 x 1            &  1 x 1        &  1 x 1     &  -           & -         \\
Pooling size           &   3 x 3             & -                   & -               &   -          &   -          &  -        \\
Pooling stride         &   2 x 2             & -                  & -                &  -           &   -          &  -        \\
Zero padding size  &   -                   & -                   & -                &  -           &   -          &  -         \\  
Spacial input size   &   61 x 61         & 14 x 14       & 10 x 10      & 8 x 8      & 6 x 6   &  1 x 1   \\ \hline
    \end{tabular}
\end{center}
\caption{Architecture details of the proposed deep poselets model.}
\label{tab:arch}
\end{table}

The output layer is a 151-way softmax: 150 different  poselet
types, plus a background class. We split the data into 4000 batches, each containing 750 positive examples (5 from each poselet type) and 750 negatives. We use a cross-entropy loss and
training for 30 epochs, annealing the learning rate by 0.2 every time
the training loss plateaus. We also used a weight decay of 1e-5 and
momentum of 0.9.

The network parameters are carefully chosen so that the network can be "unrolled" and efficiently evaluated in a grid at test time similar to OverFeat~\cite{Sermanet13}. Specifically, we approximated the mean as a constant color per channel and we designed the architecture so that each layer uses only inputs from the previous layers (and no zero padding at the edges). 

\subsubsection{Pose Discriminative Feature}

After training the net on the 150 poselet types used by~\cite{Bourdev09}, we remove the final softmax layer and used the FC-6 layer as a generic pose-discriminative feature representation. As the experiments show, this compact 256-dimensional vector is very effective for training new kinds of poselets beyond the open-sourced ones. We use it as the underlying feature vector for our poselet classifiers.

\subsection{Training the Deep Pose Representation}

\begin{figure}[h!]
\begin{center}
\vspace{-5mm}
\includegraphics[width=4.5in]{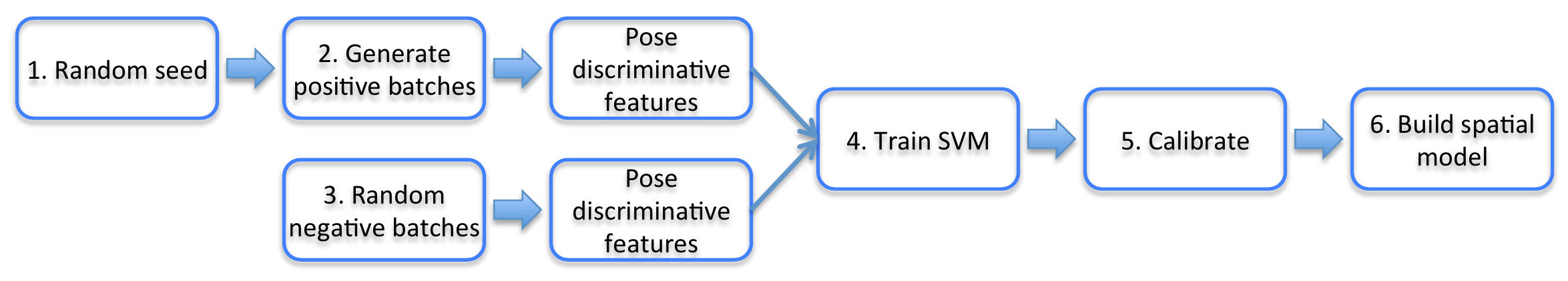}
\end{center}
\vspace*{-0.3cm}
\caption{Training Poselet Classifiers}
\label{fig:flowchart}
\vspace*{-0.3cm}
\end{figure}

We followed the algorithm of~\cite{Bourdev09} for constructing poselet classifiers. We briefly summarize it here.

Poselets are built from a set of training examples where each instance
has been annotated with 20 fiducial points on the human skeleton,
known as {\em keypoints}. Each poselet is defined by a {\em seed}
patch from a single image that captures a small subset of these
keypoints, in a precise configuration. E.g. two legs,
side-by-side, or a hand next to a hip. Given this seed (step 1), we then find other patches containing
keypoints in similar configurations in the training set by finding the optimal translation, rotation and scale that aligns the keypoints the same way as the seed patch (step 2). Crucially, the similarity does
{\em not rely on pixel appearance}, only on the distance between
constellations of keypoints. This allows us to learn sub-parts of
people in a way that is invariant to the appearance of the clothing
they are wearing, occlusions, and lighting variations.

Given a set of different seed bounding boxes, randomly sampled from the training set, their associated patches and a large set of background patches (step 3), we generate the Pose Discriminative Features of our patches and train a poselet type classifier (step 4) to discriminate that specific type from the background. Training on these patches, whose appearance varies considerably but has fixed pose, enables us to find the same body part on test examples where we have no keypoint annotations. We found that a simple linear SVM classifier on the PDF features is effective. However, it is important to "bootstrap" the training. That is, after the initial training we apply the classifier over many images not containing people, collect any false positives and retrain the class.

After training 1000 classifiers associated with randomly chosen poselets, we need to calibrate each classifier (step 6) by running them in a scanning window fashion over several hundred images and collecting the top K highest-scoring hits. We then set the threshold of each classifier to the score of the K-th highest hit. This allowed us to control the firing rate of each classifier. After calibrating the firing rate, we need to calibrate the score of each classifier. To do that, we run the classifier in a scanning fashion over the training set and detect a number of true and false positive activations. An activation is considered a true positive if it overlaps by more than 0.5 with a patch which was used as a training example for the poselet type. Once we label each activation of the poselet, and we have the activation score, we train a logistic regressor to convert the score into a probability.

At this point we have built a large library of 1000 poselet classifiers, some of which do not train well or are very similar to others. Our next step is to select a small set of the poselets, whose classifiers work well and are complementary. We use a greedy algorithm by choosing classifiers that can provide good coverage over the annotated people in our training set as described in~\cite{Bourdev09}. We collect the set of poselet classifiers and fit Gaussians to the keypoint locations associated with the positive examples of each poselet type. For example the "frontal face" poselet has a peaked Gaussian estimate of the location of the nose and eyes. In addition, each poselet is trained to predict the bounding box of the person hypothesis relative to its location.

\subsection{Training Object-Level Classifiers}

\begin{figure}[h]
\begin{center}
\includegraphics[width=4.5in]{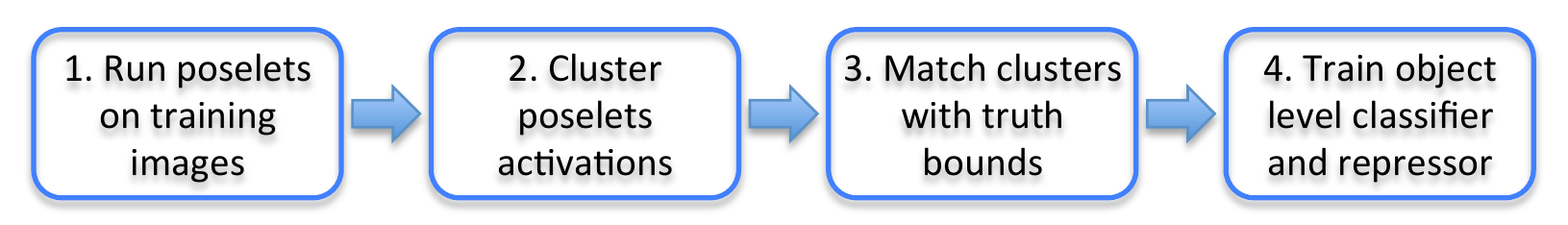}
\end{center}
\vspace*{-0.3cm}
\caption{Training Object-Level Classifiers}
\label{fig:flowchart}
\vspace*{-0.3cm}
\end{figure}

In the previous section we trained scanning classifiers for a selected set of poselet types. Here we describe how to combine them and train a classifier to predict the locations and probability of people in images.

First, we run all the poselet classifiers over the training images (Step 1) and we cluster together any activations that have compatible predictions for the keypoints of a person. Specifically we look at the symmetrized KL-divergence of the probability distributions of the keypoints predicted by two poselets and, if it is less than a threshold (set via cross-validation) we assume the activations correspond to the same person and put them in the same cluster (Step 2)

For each poselet cluster we predict the bounding box of the associated person as the consensus (mean) bounding box predicted by the poselet activations supporting the cluster. We intersect the predicted bounds with the ground truth person bounds on the training set (Step 3) and thus obtain a label for each cluster which is positive if the cluster overlaps sufficiently with a ground truth box, or negative otherwise.  We then train an SVM (Step 5) to classify the cluster into true and false positive. We also train a linear regressor to map the predicted bounds to the true bounds.

\subsection{Evaluating on a test image}

\begin{figure}[h]
\begin{center}
\includegraphics[width=4.5in]{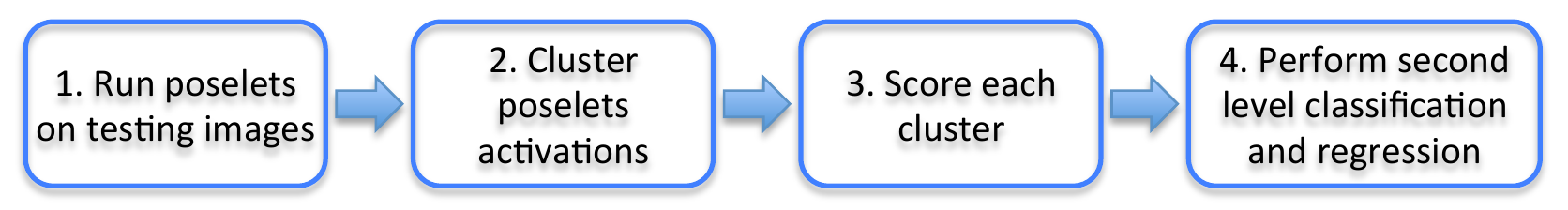}
\end{center}
\vspace*{-0.3cm}
\caption{Evaluation classifiers at test time}
\label{fig:flowchart}
\vspace*{-0.3cm}
\end{figure}

At test time we rescale the image so its largest dimension is 1000 pixels and we run the deep poselets classifier to compute the pose-discriminative feature at a comprehensive set of locations and scales. We evaluate the poselet-type SVMs on the features and collect a set of activations (step 1). We then cluster them to obtain hypothesis clusters and compute the score and regressed bounds of each cluster (step 2). We found it important to perform one final refinement of the classification score and regression bounds by applying a large, higher resolution image-net classifier and regressor. Specifically, we used the CNN classifier and bounding box regressor proposed by~\cite{Ross}. This classifier is originally trained on ImageNet but has been fine-tuned to discriminate the 20 classes in the PASCAL competition. The regressor was built on top of the Pool-5 features of the same net. We did not fine-tune the classifier and regressor which are trained to operate on region-based proposal. We suspect our performance will improve if we were to fune-tune them.

\section{Experiments}

We start by comparing the performance of our deep poselet features to
the HOG-based poselets, introduced in \cite{Bourdev09}.

\subsection{Evaluation of Feature Representation}
\label{sec:feat_eval}
\begin{figure}[h!]
\begin{center}
\includegraphics[width=4.5in]{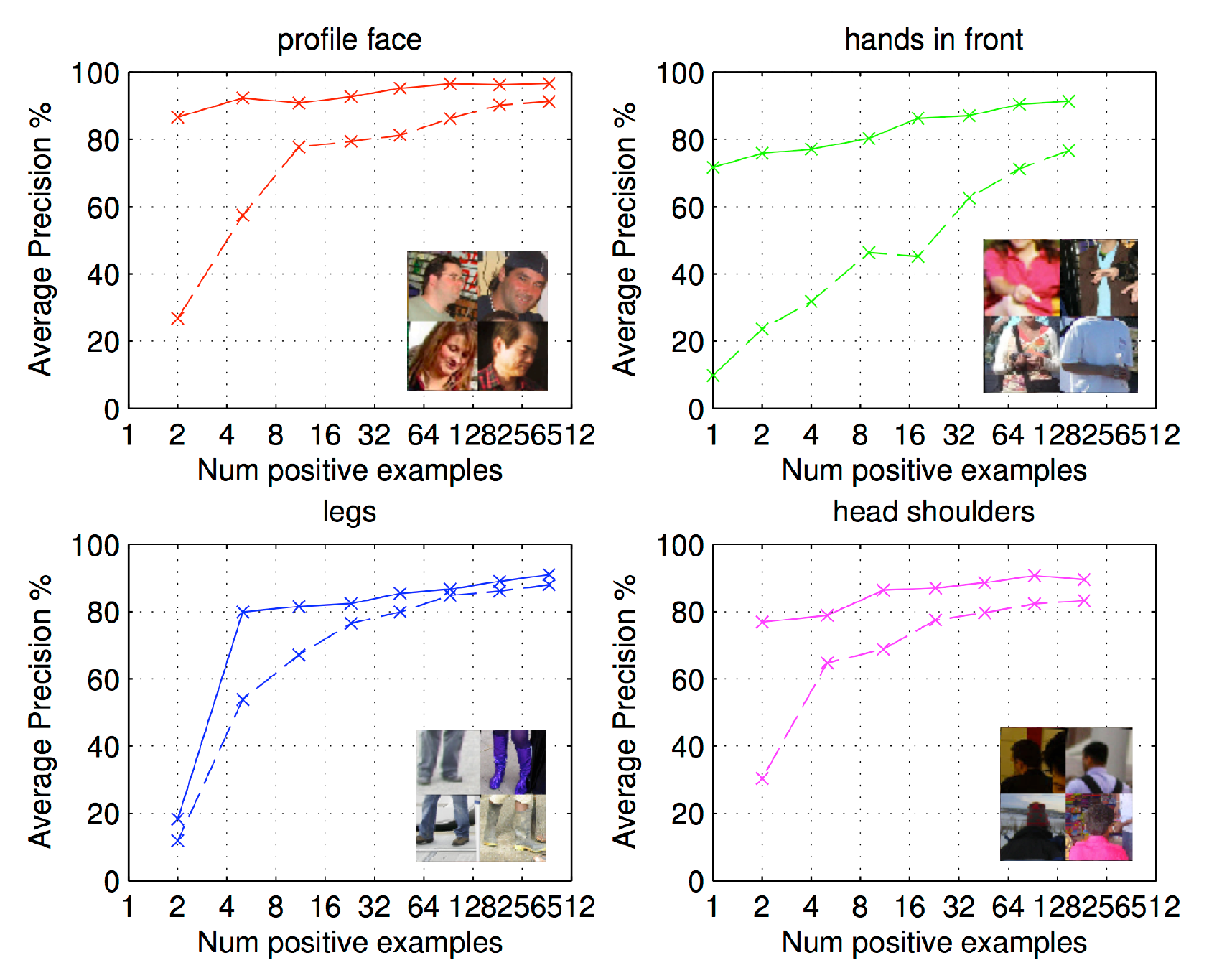}
\end{center}
\vspace*{-0.3cm}
\caption{A comparison between HOG-based poselets (dashed) and our deep
  feature-based poselets (solid), for 4 different types of
  poselet. The deep features give superior performance, particularly
  when only a few training examples are available. }
\label{fig:deep_vs_hog}
\vspace*{-0.3cm}
\end{figure}

For a given seed window we extracted up to 500 positive examples for
several poselets capturing profile face, legs, etc. The examples are
pulled from the poselets training set which includes keypoint
annotations. We then computed HOG features and deep features coming
from fc1 of our CNN. HOG features had 1476 dimensions whereas the deep
features had only 256 dimensions.  We randomized their indices and
split them into training and test sets using 25\% of the positive and
25\% of the negative examples as test and the rest as training and
trained a linear SVM as a binary classification problem (the poselet
vs the background). We computed the mean Average Precision on the test
set (which has 125 positive and 2500 negative examples). We measured
performance by training on the full training set as well as on smaller
portions of it. The results are shown in \fig{deep_vs_hog} for four
different poselets (profile face, hands in front of body, legs, head
and shoulders). 

\subsubsection{Resilience to Jittering}

We also explored the performance of our learned representation to
small misalignments that will occur at test time. We use the frontal
face poselet and generate jittered versions which have random
rotations of +/-20 degrees, scalings of 0.7 to 1.3 and translations of
+/- 16 pixels (out of 64) in both x and y dimensions. The ranges are sampled
uniformly for each poselet example. \tab{jitter} shows the performance
of the HOG and learned representations as the number of training
examples is varied.


\begin{minipage}{\textwidth}
  \begin{minipage}[b]{0.35\textwidth}
    \centering
\includegraphics[width=1.5in]{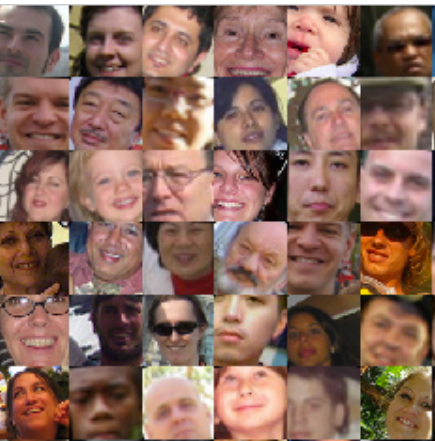}
    \captionof{figure}{Some jittered examples.}
  \end{minipage}
  \hfill
\vspace{-3mm}
  \begin{minipage}[b]{0.60\textwidth}
    \centering
    \small
\vspace{-3mm}
    \begin{tabular}{| c | c | c | c |}
    \hline
Pos. & Neg. & HOG AP \% & Deep AP \% \\ \hline
375 & 7500 & 70.59 & 99.44 \\ \hline 
187 & 3750 & 64.87 & 98.92 \\ \hline
93 & 1875 & 59.1 2 & 98.47 \\ \hline 
46 & 937 & 55.79 &  97.13 \\ \hline
23 & 468 & 48.84 & 95.22  \\ \hline
11 & 234 & 45.6 & 92.94 \\ \hline 
5 & 117 & 32.45 & 88.35 \\ \hline
2 & 58 & 29.83 & 82.58 \\
    \hline
\end{tabular}
\label{tab:jitter}
      \captionof{table}{Comparison between HOG-based and deep feature-based poselets
  for a jittered test set. }
    \end{minipage}
  \end{minipage}
\vspace{3mm}



In summary, our learned representation:
\begin{itemize}
\item Gives significantly higher accuracy than HOG.
\item Perform very impressively with few training examples. This is
  particularly important because some poselets are very rare and we
  cannot train HOG-based models on them.
\item Is more compact (1476 vs 256 dimensions) and therefore SVM evaluation is faster
\item Is more resilient to misalignment of the data. This is important
  because it implies we can run them at coarser step and the system
  could run much faster. However, they may be less precise than HOG in
  predicting the keypoints of the detected person.
\end{itemize}

Note that here we did not try fine-tuning the poselet network, instead
just training a linear SVM on the fixed final-layer
features. Furthermore, the learned features were also trained on a different
dataset to recognize different sets of poselets, albeit of very
similar patterns.

\subsection{PASCAL Evaluation}

We evaluate our algorithm of the PASCAL VOC datasets \cite{PASCAL}, which show
people in a wide range of natural settings. The large variation in
scale, occlusion and pose make the dataset highly challenging. 
These datasets have also been used to evaluate many of the leading
approaches. In particular, the recent work of Girshick \etal
\cite{Ross} which also make used of convolutional networks, has the
current leading performance on the person detection task with a score
of 58.0\% average precision. As shown in \tab{person_ap}, our
algorithm outperforms them by 0.6-1.2\%. 

\begin{table}[h]
\begin{center}
    \begin{tabular}{| c | c | c | c |}
    \hline
Test set  &  HOG Poselets & RCNN & Deep Poselets        \\ \hline
VOC 2007  & 46.9 & 58.7  &  \textbf{59.3}     \\
VOC 2010  & 48.5 & 58.1  &  \textbf{59.3}     \\
VOC 2011  & - & 57.8  &  \textbf{58.7}      \\ \hline
    \end{tabular}
\end{center}
\caption{mAP of person detection compared with R-CNN.}
\label{tab:person_ap}
\end{table}

Note that we achieved this high performance even though our poselet model is significantly simplified. In~\tab{differences_with_hog} we show the differences of our method vs the original poselets of~\cite{BourdevPoseletsECCV10}. The original poselets have various enhancements -- multiple aspect ratios, contextual poselets, dynamic selection of the number of training examples per poselet, etc. Each of these choices adds about 1-2\% to the accuracy but at the expense of making the system more complicated and slower to train and evaluate. We have decided to disable these extensions favoring simplicity over higher performance. We used a fixed square aspect ratio and our patch resolution is on average 40\% smaller. For object level classification and regression we use the pre-trained CNN of~\cite{Ross}. In this work we use the CNN without fine-tuning it for our task.

\begin{table}[h]
\begin{center}
    \begin{tabular}{| c | c | c |}
    \hline
  &  HOG Poselets~\cite{BourdevPoseletsECCV10} & Deep Poselets        \\ \hline
Poselet classifier features & HOG & PDF \\
Pixels per patch & 4096 - 8192 & 3721 \\
Aspect ratios & 64x96 64x64 96x64 128x64 & 61x61 \\
Contextual poselets (Q-poselets) & enabled & not used \\
num training examples per poselet & dynamic 200 - 1000+ & 200 \\ 
Object-level classification features & poselet scores & FC7 layer of R-CNN~\cite{Ross} \\
Object-level regression features & consensus bounds & Pool5 layer of R-CNN~\cite{Ross}\\ \hline
    \end{tabular}
\end{center}
\caption{Detailed differences from the HOG poselets of~\cite{BourdevPoseletsECCV10}.}
\label{tab:differences_with_hog}
\vspace{-3mm}
\end{table}

\begin{figure}[h!]
 \begin{center}
\includegraphics[width=2.32in]{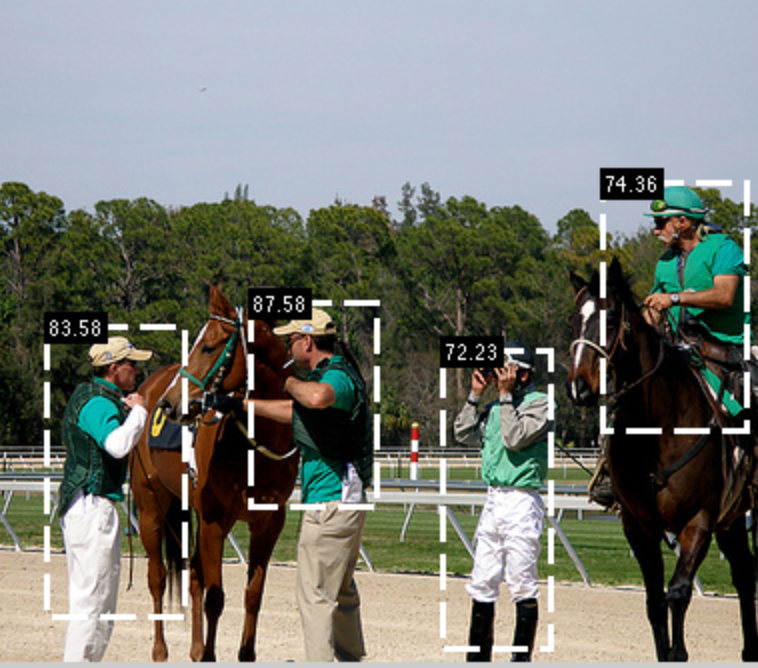}
\includegraphics[width=2.5in]{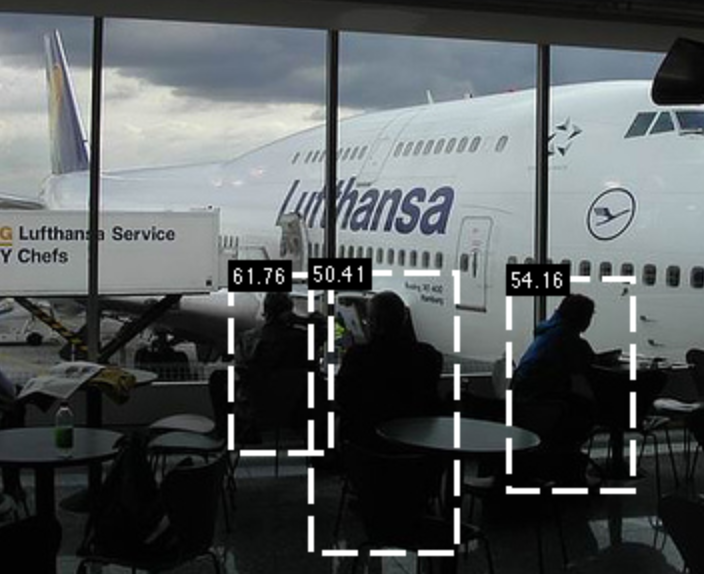}
\end{center}
 \vspace*{-0.3cm}
 \caption{Some detection examples. }
 \label{fig:ex}
 \vspace*{-0.3cm}
 \end{figure}

\vspace{-3mm}
\section{Conclusion}
\vspace{-2mm}
We described a pipeline that combines deep features, explicitly
trained to discriminate between different poses with the poselets
framework of \cite{Bourdev09} to generate high quality
bounding box proposals for people. Using the RCNN person detector of
\cite{Ross}, we then score these proposals and refine the bounding box
coordinates. The system produces state-of-the-art performance, beating the RCNN
detector alone by 0.6--1.2\% on the highly challenging PASCAL
datasets. Surpassing the RCNN detector is significant,
given that it gave a $\sim$10\% AP gain over the previous state-of-the-art.
The results show the importance of maintaining the correct aspect
ratio when classifying regions and give further evidence for the power
of deep features over hand-crafted features such as HOG.

\vspace{-3mm}
{
\small
\bibliographystyle{ieee}
\bibliography{deep_poselets}

\begin{thebibliography}{10}\itemsep=-1pt

\bibitem{taigman14}
{DeepFace: Closing the Gap to Human-Level Performance in Face Verification}.
\newblock In {\em Conference on Computer Vision and Pattern Recognition
  (CVPR)}.

\bibitem{panda}
{PANDA: Pose Aligned Networks for Deep Attribute Modeling}.
\newblock In {\em Conference on Computer Vision and Pattern Recognition
  (CVPR)}.

\bibitem{PASCAL}
The pascal visual object classes challenge.
\newblock \url{http://pascallin.ecs.soton.ac.uk/challenges/VOC/}.

\bibitem{BourdevPoseletsECCV10}
L.~Bourdev, S.~Maji, T.~Brox, and J.~Malik.
\newblock Detecting people using mutually consistent poselet activations.
\newblock In {\em European Conference on Computer Vision (ECCV)}, 2010.

\bibitem{Bourdev09}
L.~Bourdev and J.~Malik.
\newblock Poselets: Body part detectors trained using {3D} human pose
  annotations.
\newblock In {\em International Conference on Computer Vision (ICCV)}, 2009.

\bibitem{DalalTriggs}
N.~Dalal and B.~Triggs.
\newblock Histograms of oriented gradients for human detection.
\newblock In C.~Schmid, S.~Soatto, and C.~Tomasi, editors, {\em International
  Conference on Computer Vision \& Pattern Recognition}, volume~2, pages
  886--893, INRIA Rh\^one-Alpes, ZIRST-655, av. de l'Europe, Montbonnot-38334,
  June 2005.

\bibitem{dollar2010fastest}
P.~Doll{\'a}r, S.~Belongie, and P.~Perona.
\newblock The fastest pedestrian detector in the west.
\newblock {\em BMVC}, 2(3):7, 2010.

\bibitem{Farabet12}
C.~Farabet, C.~Couprie, L.~Najman, and Y.~LeCun.
\newblock Learning hierarchical features for scene labeling.
\newblock {\em Pattern Analysis and Machine Intelligence, IEEE Transactions
  on}, 35(8):1915--1929, 2013.

\bibitem{dpm}
P.~F. Felzenszwalb, R.~B. Girshick, D.~McAllester, and D.~Ramanan.
\newblock Object detection with discriminatively trained part based models.
\newblock {\em IEEE Transactions on Pattern Analysis and Machine Intelligence},
  32(9):1627--1645, 2010.

\bibitem{Fergus03}
R.~Fergus, P.~Perona, and A.~Zisserman.
\newblock Object class recognition by unsupervised scale-invariant learning.
\newblock In {\em Proceedings of the IEEE Conference on Computer Vision and
  Pattern Recognition}, volume~2, pages 264--271, June 2003.

\bibitem{Fidler13}
S.~Fidler, R.~Mottaghi, A.~Yuille, and R.~Urtasun.
\newblock Bottom-up segmentation for top-down detection.
\newblock In {\em CVPR}, 2013.

\bibitem{Ross}
R.~Girshick, J.~Donahue, T.~Darrell, and J.~Malik.
\newblock Rich feature hierarchies for accurate object detection and semantic
  segmentation.
\newblock arXiv preprint:1311.2524.

\bibitem{Krizhevsky12}
A.~Krizhevsky, I.~Sutskever, and G.~E. Hinton.
\newblock Imagenet classification with deep convolutional neural networks.
\newblock In {\em Neural Information Processing Systems (NIPS)}, 2012.

\bibitem{lecun}
Y.~LeCun, B.~Boser, J.~S. Denker, D.~Henderson, R.~E. Howard, W.~Hubbard, and
  L.~D. Jackel.
\newblock Handwritten digit recognition with a back-propagation network.
\newblock In D.~Touretzky, editor, {\em Advances in Neural Information
  Processing Systems (NIPS 1989)}, volume~2, Denver, CO, 1990. Morgan Kaufman.

\bibitem{leibe2004combined}
B.~Leibe, A.~Leonardis, and B.~Schiele.
\newblock Combined object categorization and segmentation with an implicit
  shape model.
\newblock {\em Workshop on statistical learning in computer vision, ECCV},
  2(5):7, 2004.

\bibitem{Josef}
M.~Oquab, I.~Laptev, L.~Bottou, and J.~Sivic.
\newblock Learning and transferring mid-level image representations using
  convolutional neural networks.
\newblock In {\em Computer Vision and Pattern Recognition (CVPR)}, 2014.

\bibitem{Sermanet13}
P.~Sermanet, D.~Eigen, X.~Zhang, M.~Mathieu, R.~Fergus, and Y.~LeCun.
\newblock Overfeat: Integrated recognition, localization and detection using
  convolutional networks.
\newblock In {\em International Conference on Learning Representations (ICLR)},
  April 2014.

\bibitem{sermanet-cvpr-13}
P.~Sermanet, K.~Kavukcuoglu, S.~Chintala, and Y.~LeCun.
\newblock Pedestrian detection with unsupervised multi-stage feature learning.
\newblock In {\em Proc. International Conference on Computer Vision and Pattern
  Recognition (CVPR'13)}. IEEE, June 2013.

\bibitem{Szegedy13}
C.~Szegedy, A.~Toshev, and D.~Ehran.
\newblock Deep neural networks for object detection.
\newblock 2013.

\bibitem{Uijlings}
J.~Uijlings, K.~van~de Sande, T.~Gevers, and A.~Smeulders.
\newblock Selective search for object recognition.
\newblock {\em IJCV}, 2013.

\bibitem{Viola2001}
P.~Viola and M.~Jones.
\newblock {Rapid object detection using a boosted cascade of simple features}.
\newblock {\em CVPR}, 1:I--511--I--518, 2001.

\bibitem{wang}
X.~Wang, M.~Yang, S.~Zhu, and Y.~Lin.
\newblock Regionlets for generic object detection.
\newblock In {\em ICCV}, 2013.

\bibitem{Matt}
M.~D. Zeiler and R.~Fergus.
\newblock Visualizing and understanding convolutional neural networks.
\newblock arXiv preprint:1311.2901.

\end{thebibliography}
}

\end{document}